\documentclass{ipmu2008}

\usepackage{
latexsym,
graphicx,
times,
amsmath,
amssymb,
psfrag,
subfigure}

\newtheorem{theorem}{Theorem}

\title{Cumulative and Averaging Fission of Beliefs}

\author{{\bf Audun J{\o}sang}\\
UniK, University of Oslo \\
Norway\\
josang~@~unik.no}

\begin{document}
\maketitle
\begin{abstract}
Belief fusion is the principle of combining separate beliefs or bodies
of evidence originating from different sources. Depending on the
situation to be modelled, different belief fusion methods can be
applied. Cumulative and averaging belief fusion is defined for fusing
opinions in subjective logic, and for fusing belief functions in
general. The principle of fission is the opposite of fusion, namely to
eliminate the contribution of a specific belief from an already fused
belief, with the purpose of deriving the remaining belief. This paper
describes fission of cumulative belief as well as fission of averaging
belief in subjective logic. These operators can for example be applied
to belief revision in Bayesian belief networks, where the belief
contribution of a given evidence source can be determined as a
function of a given fused belief and its other contributing
beliefs. 


\end{abstract}

\section{Introduction}

Subjective logic is a type of probabilistic logic that explicitly
takes uncertainty and belief ownership into account. In general,
subjective logic is suitable for modeling and analysing situations
involving uncertainty and incomplete knowledge
\cite{Jos1997-AWCR,Jos2001-IJUFKS}. For example, it can be used for
modeling trust networks \cite{JGK2006-WIAS} and for analysing Bayesian
networks \cite{Jos2008-JMVLSC}.

Arguments in subjective logic are subjective opinions about
propositions. The opinion space is a subset of the belief function
space used in Dempster-Shafer belief theory. The term belief will be
used interchangeably with opinions throughout this paper. A binomial opinion applies to a
single proposition, and can be represented as a Beta distribution. A
multinomial opinion applies to a collection of propositions, and can
be represented as a Dirichlet distribution. Through the correspondence
between opinions and Beta/Dirichlet distributions, subjective logic
provides an algebra for these functions.

The two types of fusion defined for subjective logic are {\em
cumulative fusion} and {\em averaging fusion}
\cite{Jos2007-CATS}. Situations that can be modelled with the
cumulative operator are for example when fusing beliefs of two
observers who have assessed separate and independent evidence, such as
when they have observed the outcomes of a given process over two
separate non-overlapping time periods. Situations that can be modelled
with the averaging operator are for example when fusing beliefs of two
observers who have assessed the same evidence and possibly interpreted
it differently.

Dempster's rule also represents a method commonly applied for fusing
beliefs. However, it is not used in subjective logic and will not be
discussed here.

There are situations where it is useful to split a fused belief in
its contributing belief components, and this process is called belief
fission. This requires the already fused belief and one of its contributing
belief components as input, and will produce the remaining
contributing belief component as output. Fission is basically the
opposite of fusion, and the formal expressions for fission can be
derived by rearranging the expressions for fusion. This will be
described in the following sections.

\section{Fundamentals of Subjective Logic}

Subjective opinions express subjective beliefs about the truth of
propositions with degrees of uncertainty, and can indicate subjective
belief ownership whenever required. An opinion is usually denoted as
$\omega^{A}_{x}$ where $A$ is the subject, also called the belief
owner, and $x$ is the proposition to which the opinion applies. An
alternative notation is $\omega(A\!:\!x)$. The proposition $x$ is
assumed to belong to a frame of discernment (also called state space)
e.g. denoted as $X$, but the frame is usually not included in the
opinion notation. The propositions of a frame are normally assumed to
be exhaustive and mutually disjoint, and subjects are assumed to have
a common semantic interpretation of propositions. The subject, the
proposition and its frame are attributes of an opinion. Indication of
subjective belief ownership is normally omitted whenever irrelevant.

\subsection{Binomial Opinions}

Let $x$ be a proposition. Entity $A$'s binomial opinion about the
truth of a $x$ is the ordered quadruple $\omega^{A}_{x} =
(b,d,u,a)$ with the components:
\vspace{2ex}\\
\begin{tabular}{ll}
$b$: &belief that the proposition is true\\
$d$: &disbelief that the proposition is true\\
     &(i.e. the belief that the proposition is false)\\
$u$: &uncertainty about the probability of $x$\\
     &(i.e. the amount of uncommitted belief)\\
$a$: &base rate of $x$\\
     &(i.e. probability of $x$ in the absence of belief)
\end{tabular}

These components satisfy:
\begin{eqnarray}
\label{eq:scalar-constraint}
b,\;d,\;u,\;a \in [0,1]\\
\nonumber\\
\label{eq:additivity-constraint}
\mbox{and }\;\;\;\;\;\;\;\;\;\;
b + d + u = 1
\end{eqnarray}

The characteristics of various opinion classes are listed below.
An opinion where:\\
\begin{tabular}{l}
$b=1$: is equivalent to binary logic TRUE,\\
$d=1$: is equivalent to binary logic FALSE,\\
$b+d=1$: is equivalent to a probability,\\
$0<(b+d)<1$: expresses uncertainty, and\\
$b+d=0$: is vacuous (i.e. totally uncertain).
\end{tabular}

The probability expectation value of a binomial opinion is:
\begin{equation}
\label{eq:expectation}
p(\omega_{x}) = b_{x} + a_{x}u_{x}\;.
\end{equation}
The expression of Eq.(\ref{eq:expectation}) is equivalent to the
pignistic probability in traditional belief function theory
\cite{SK1994}, and is based on the principle that the belief mass
assigned to the whole frame is split equally among the singletons of
the frame. In Eq.(\ref{eq:expectation}) the base rate $a_{x}$ must be
interpreted in the sense that the relative proportion of singletons
contained in $x$ is equal to $a_{x}$.

Binomial opinions can be represented on an equilateral triangle as
shown in Fig.\ref{fig:tri-proj-2} below. A point inside the triangle
represents a $(b,d,u)$ triple. The b,d,u-axes run from one edge to the
opposite vertex indicated by the Belief, Disbelief or Uncertainty
label. For example, a strong positive opinion is represented by a
point towards the bottom right Belief vertex. The base rate, also
called relative atomicity, is shown as a red pointer along the
probability base line, and the probability expectation, E, is formed
by projecting the opinion onto the base, parallel to the base rate
projector line. As an example, the opinion
$\omega_{x}=(0.4,\;0.1,\;0.5,\;0.6)$ is shown on the figure.

\begin{figure}[h]
\begin{center}
\includegraphics[scale=0.45]{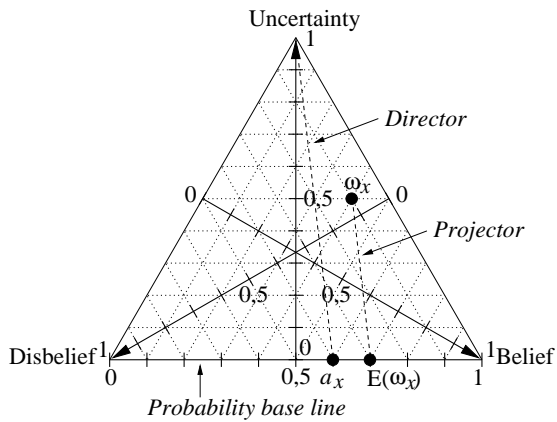}
\caption{Opinion triangle with example opinion}
\label{fig:tri-proj-2}
\end{center}
\end{figure}

Uncertainty about probability values can be interpreted as ignorance,
or second order uncertainty about the first order probabilities. In
this paper, the term ``uncertainty'' will be used in the sense
of {\em ``uncertainty about the probability values''}. A probabilistic
logic based on belief theory therefore represents a generalisation of
traditional probabilistic logic.

\subsection{Multinomial Opinions}

Let $X$ be a frame, i.e. a set of exhaustive and mutually disjoint
propositions $x_{i}$. Entity $A$'s multinomial opinion over $X$ is the
composite function $\omega^{A}_{X}=(\vec{b}, u, \vec{a})$, where
$\vec{b}$ is a vector of belief masses over the propositions of
$X$, $u$ is the uncertainty mass, and $\vec{a}$ is a vector of
base rate values over the propositions of $X$. These components
satisfy:
\begin{eqnarray}
\label{eq:multi-scalar-constraint}
\vec{b}(x_{i}),u,\vec{a}(x_{i}) \in [0,1],\;\; \forall x_{i} \in X\\
\nonumber\\
\label{eq:multi-additivity-constraint}
u+\sum_{x_{i} \in X} \vec{b}(x_{i}) = 1\\
\nonumber\\
\label{eq:base-rate-additivity-constraint}
\sum_{x_{i} \in X} \vec{a}(x_{i}) = 1
\end{eqnarray}

\begin{figure}[h]
\begin{center}
\includegraphics[scale=0.45]{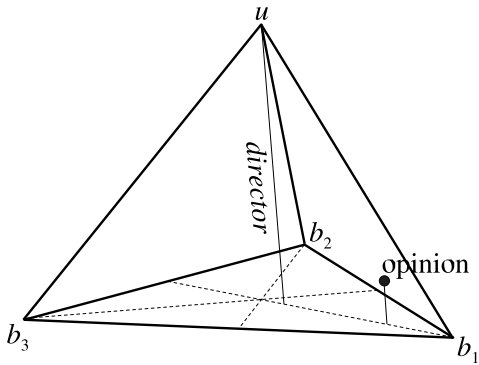}
\caption{Opinion pyramid with example trinomial opinion}
\label{fig:pyramid-simple}
\end{center}
\end{figure}

Visualising multinomial opinions is not trivial. Trinomial opinions
can be visualised as points inside a triangular pyramid as shown in
Fig.\ref{fig:pyramid-simple}, but the 2D aspect of printed paper and
computer monitors make this impractical in general.

Opinions with
dimensions larger than trinomial do not lend themselves to traditional
visualisation.

\section{Fusion of Multinomial Opinions}
In many situations there will be multiple sources of evidence, and
fusion can be used to combine evidence from different sources.

In order to provide an interpretation of fusion in subjective logic it
is useful to consider a process that is observed by two
sensors. A distinction can be made between two cases.
\begin{enumerate}
\item The two sensors observe the process during disjoint time
periods. In this case the observations are independent, and it is
natural to simply add the observations from the two sensors, and the
resulting fusion is called {\em cumulative fusion}.

\item The two sensors observe the process during the same time
period. In this case the observations are dependent, and it is natural
to take the average of the observations by the two sensors, and the
resulting fusion is called {\em averaging fusion}.
\end{enumerate}

\subsection{Cumulative Fusion}

Assume a frame $X$ containing $k$ elements.  Assume two observers $A$
and $B$ who have independent opinions over the frame $X$. This cane
for example result from having observed the outcomes of a process
over two separate time periods.

Let the two observers' respective opinions be expressed as
$\omega^{A}_{X} = (\vec{b}^{A}_{X}, u^{A}_{X}, \vec{a}^{A}_{X})$ and
$\omega^{B}_{X} = (\vec{b}^{B}_{X}, u^{B}_{X}, \vec{a}^{B}_{X})$.

The cumulative fusion of these two bodies of evidence is denoted as
$\omega{A\diamond B}_{X} = \omega^{A}_{X} \oplus \omega^{B}_{X}$. The
symbol ``$\diamond$'' denotes the fusion of two observers $A$ and $B$
into a single imaginary observer denoted as $A \diamond B$. The
mathematical expressions for cumulative fusion is described below.

\begin{theorem}{\bf The Cumulative Fusion Operator}\\
\label{theorem:cumulative-fusion}
{  {
\hspace{1ex}\\ Let $\omega^{A}_{X}$ and $\omega^{B}_{X}$ be opinions
respectively held by agents $A$ and $B$ over the same frame
$X = \{x_i\;|\; i =1,\cdots,k\}$.
Let $\omega^{A\diamond B}_{X} $ be the opinion such that:
\begin{eqnarray}
\mbox{Case I:}\;\;\; \mbox{For }\;\; u^{A}_{X} \neq 0\;\; \lor\;\;  u^{B}_{X} \neq 0 :
\nonumber \\
\hline
\nonumber \\
\left\{
\begin{array}{ll}
b^{A\diamond B}_{x_{i}}&= \frac{b^{A}_{x_{i}}u^{B}_{X} 
+ b^{B}_{x_{i}}u^{A}_{X}}
{u^{A}_{X} + u^{B}_{X} - u^{A}_{X}u^{B}_{X}}\\\\
u^{A\diamond B}_{X}     &=  \frac{u^{A}_{X}u^{B}_{X}}{u^{A}_{X} + u^{B}_{X} - u^{A}_{X}u^{B}_{X}}
\end{array}
\right.
\label{eq:cumulative-fusion}
\end{eqnarray}

\begin{eqnarray}
\mbox{Case II:}\;\;\;\mbox{For }\;\; u^{A}_{X} = 0\;\; \land\;\; u^{B}_{X} = 0:
\nonumber \\
\hline
\nonumber \\
\left\{
\begin{array}{ll}
b^{A\diamond B}_{x_{i}}&= \gamma \, b^{A}_{x_{i}} + (1-\gamma)b^{B}_{x_{i}}\\\\
u^{A\diamond B}_{X} &= 0
\end{array}
\right.
\label{eq:zero-div-cumulative-fusion}\\
\nonumber\\
\nonumber
\mbox{where }\;\;
\gamma = \lim \limits_{\substack{ 
{u^{A}_{X} \rightarrow 0}\\
{u^{B}_{X} \rightarrow 0}
}}
\;\;\frac{u^{B}_{X}}{u^{A}_{X} + u^{B}_{X}}
\end{eqnarray}

Then $\omega^{A\diamond B}_{X}$ is called the cumulatively fused opinion of
$\omega^{A}_{X}$ and $\omega^{B}_{X}$, representing the combination of
independent opinions of $A$ and $B$. By using the symbol `$\oplus$' to
designate this belief operator, we define $\omega^{A\diamond B}_{X} \equiv
\omega^{A}_{X} \oplus \omega^{B}_{X}$.  }  }
\end{theorem}

The cumulative fusion operator is equivalent to {\em a posteriori}
updating of Dirichlet distributions. Its proof and derivation is based
on the bijective mapping between multinomial opinions and and an
augmented representation of the Dirichlet distribution
\cite{Jos2007-CATS}.

It can be verified that the cumulative fusion operator is commutative,
associative and non-idempotent. In Case II of
Theorem~\ref{theorem:cumulative-fusion}, the associativity depends on
the preservation of relative weights of intermediate results, which
requires the additional weight variable $\gamma$. In this case, the
cumulative operator is equivalent to the weighted average of
probabilities.

The cumulative fusion operator represents a generalisation of the consensus
operator \cite{Jos2002-AIJ,Jos2001-IJUFKS} which
emerges directly from Theorem~\ref{theorem:cumulative-fusion} by
assuming a binary frame.

\subsection{Averaging Fusion}

Assume a frame $X$ containing $k$ elements.  Assume two observers $A$
and $B$ who have dependent opinions over the frame $X$. This can for
example result from observing the outcomes of the process over the
same time periods.

Let the two observers' respective opinions be expressed as
$\omega^{A}_{X} = (\vec{b}^{A}_{X}, u^{A}_{X}, \vec{a}^{A}_{X})$ and $\omega^{B}_{X} = (\vec{b}^{B}_{X}, u^{B}_{X}, \vec{a}^{B}_{X})$.

The averaging fusion of these two bodies of evidence is denoted as
$\omega{A\underline{\diamond} B}_{X} = \omega^{A}_{X}
\underline{\oplus} \omega^{B}_{X}$. The symbol
``$\underline{\diamond}$'' denotes the averaging fusion of two
observers $A$ and $B$ into a single imaginary observer denoted as $A
\underline{\diamond} B$. The mathematical expressions for averaging fusion is
described below.

\begin{theorem}{\bf The Averaging Fusion Operator}\\
\label{theorem:averaging-fusion}
\hspace{1ex}\\ Let $\omega^{A}_{X}$ and $\omega^{B}_{X}$ be opinions
respectively held by agents $A$ and $B$ over the same frame
$X = \{x_i\;|\; i =1,\cdots,k\}$.
Let $\omega^{A\underline{\diamond} B}_{X}$ be the opinion such that:

\begin{eqnarray}
\mbox{Case I:}\;\;\; \mbox{For }\;\; u^{A}_{X} \neq 0\;\; \lor\;\;  u^{B}_{X} \neq 0 :
\nonumber \\
\hline
\nonumber \\
\left\{
\begin{array}{ll}
b^{A\underline{\diamond} B}_{x_{i}}&= 
\frac{
b^{A}_{x_{i}}u^{B}_{X} + 
b^{B}_{x_{i}}u^{A}_{X}}
{u^{A}_{X} + u^{B}_{X}}\\\\
u^{A\underline{\diamond} B}_{X} &= \frac{2u^{A}_{X}u^{B}_{X}}{u^{A}_{X} + u^{B}_{X}}
\end{array}
\right.
\label{eq:averaging-fusion}
\end{eqnarray}

\begin{eqnarray}
\mbox{Case II:}\;\;\; \mbox{For }\;\; u^{A}_{X} = 0\;\; \land\;\;  u^{B}_{X} = 0:
\nonumber \\
\hline
\nonumber \\
\left\{
\begin{array}{ll}
b^{A\diamond B}_{x_{i}}&= \gamma \, b^{A}_{x_{i}} + (1-\gamma) b^{B}_{x_{i}}\\\\
u^{A\diamond B}_{X} &= 0
\end{array}
\right.
\label{eq:zero-div-averaging-fusion}\\
\nonumber\\
\nonumber
\mbox{where }\;\;
\gamma = \lim \limits_{\substack{ 
{u^{A}_{X} \rightarrow 0}\\
{u^{B}_{X} \rightarrow 0}
}}
\;\;\frac{u^{B}_{X}}{u^{A}_{X} + u^{B}_{X}}
\end{eqnarray}

Then $\omega^{A\underline{\diamond} B}_{X}$ is called the averaged opinion
of $\omega^{A}_{X}$ and $\omega^{B}_{X}$, representing the combination of the
dependent opinions of $A$ and $B$. By using the symbol
`$\underline{\oplus}$' to designate this belief operator, we define
$\omega^{A\underline{\diamond} B}_{X} \equiv \omega^{A}_{X} \underline{\oplus}
\omega^{B}_{X}$.
\end{theorem}

The averaging operator is equivalent to averaging the evidence of
Dirichlet distributions. Its proof derivation is based on the
bijective mapping between multinomial opinions and an augmented
representation of Dirichlet distributions \cite{Jos2007-CATS}.

It can be verified that the averaging fusion operator is commutative
and idempotent, but not associative.

The averaging fusion operator represents a generalisation of the
consensus operator for dependent opinions defined in
\cite{JK1998-NISSC}.

\section{Fission of Multinomial Opinions}

The principle of belief fission is the opposite to belief fusion. This
section describes the fission operators corresponding to the cumulative
and averaging fusion operators described in the previous section.

\subsection{Cumulative Fission}

Assume a frame $X$ containing $k$ elements.  Assume two observers $A$
and $B$ who have observed the outcomes of a process over two separate
time periods. Assume that the observers beliefs have been cumulatively
fused into $\omega^{A\diamond B}_{X} = \omega^{C}_{X} = (\vec{b}^{C}_{X},
u^{C}_{X}, \vec{a}^{C}_{X})$, and assume that entity $B$'s contributing opinion $\omega^{B}_{X} = (\vec{b}^{B}_{X}, u^{B}_{X}, \vec{a}^{B}_{X})$ is known.

The cumulative fission of these two bodies of evidence is denoted as
$\omega^{C\overline{\diamond} B}_{X} = \omega^{A}_{X} = \omega^{C}_{X}
\ominus \omega^{B}_{X}$, which represents entity $A$'s contributing
opinion. The mathematical expressions for cumulative fission is
described below.

\begin{theorem}{\bf The Cumulative Fission Operator}\\
\label{theorem:cumulative-fission}
\hspace{1ex}\\ Let $\omega^{C}_{X} = \omega^{A\diamond B}_{X}$ be the
cumulatively fused opinion of $\omega^{B}_{X}$ and the unknown opinion
$\omega^{A}_{X}$ over the frame $X = \{x_i\;|\; i =1,\cdots,k\}$.  Let
$\omega^{A}_{X} = \omega^{C\overline{\diamond} B}_{X} $ be the opinion
such that:
\begin{eqnarray}
\mbox{Case I:}\;\;\; \mbox{For }\;\; u^{C}_{X} \neq 0\;\; \lor\;\;  u^{B}_{X} \neq 0 :
\nonumber \\
\hline
\nonumber \\
\left\{
\begin{array}{ll}
b^{A}_{x_{i}} = b^{C\overline{\diamond} B}_{x_{i}}
&= \frac{b^{C}_{x_{i}}u^{B}_{X} - b^{B}_{x_{i}}u^{C}_{X}}
{u^{B}_{X} - u^{C}_{X} + u^{B}_{X}u^{C}_{X}}\\\\
u^{A}_{X} = u^{C\overline{\diamond} B}_{X}     
&=  \frac{u^{B}_{X}u^{C}_{X}}{u^{B}_{X} - u^{C}_{X} + u^{B}_{X}u^{C}_{X}}
\end{array}
\right.
\label{eq:cumulative-fission}
\end{eqnarray}

\begin{eqnarray}
\mbox{Case II:}\;\;\;\mbox{For }\;\; u^{C}_{X} = 0\;\; \land\;\; u^{B}_{X} = 0:
\nonumber \\
\hline
\nonumber \\
\left\{
\begin{array}{ll}
b^{A}_{x_{i}} = b^{C\overline{\diamond} B}_{x_{i}}&= \gamma^{B} \, b^{C}_{x_{i}} - \gamma^{C} b^{B}_{x_{i}}\\\\
u^{A}_{X} = u^{C\overline{\diamond} B}_{X} &= 0
\end{array}
\right.
\label{eq:zero-div-cumulative-fission}\\
\nonumber\\
\nonumber
\mbox{where }
\left\{
\begin{array}{l}
\gamma^{B} = \lim \limits_{\substack{ 
{u^{C}_{X} \rightarrow 0}\\
{u^{B}_{X} \rightarrow 0}
}}
\;\;\frac{u^{B}_{X}}{u^{B}_{X} - u^{C}_{X} + u^{B}_{X}u^{C}_{X}}\\
\gamma^{C} = \lim \limits_{\substack{ 
{u^{C}_{X} \rightarrow 0}\\
{u^{B}_{X} \rightarrow 0}
}}
\;\;\frac{u^{C}_{X}}{u^{B}_{X} - u^{C}_{X} + u^{B}_{X}u^{C}_{X}}\\
\end{array}
\right.
\end{eqnarray}

Then $\omega^{C\overline{\diamond} B}_{X}$ is called the cumulatively
fissioned opinion of $\omega^{C}_{X}$ and $\omega^{B}_{X}$,
representing the result of eliminating the opinions of $B$ from that of
$C$. By using the symbol `$\ominus$' to designate this belief operator,
we define $\omega^{C\overline{\diamond} B}_{X} \equiv \omega^{C}_{X} \ominus
\omega^{B}_{X}$.
\end{theorem}

Cumulative fission is the inverse of cumulative fusion. Its proof and
derivation is based on rearranging the mathematical expressions of
Theorem~\ref{theorem:cumulative-fusion}

It can be verified that the cumulative rule is non-commutative,
non-associative and non-idempotent. In Case II of
Theorem~\ref{theorem:cumulative-fission},  the
fission rule is equivalent to the weighted subtraction of
probabilities.

\subsection{Averaging Fission}

Assume a frame $X$ containing $k$ elements.  Assume two observers $A$
and $B$ who have observed the same outcomes of a process over the same
time period. Assume that the observers beliefs have been averagely
fused into $\omega^{C}_{X} = \omega^{A\underline{\diamond} B}_{X} = 
(\vec{b}^{C}_{X}, u^{C}_{X}, \vec{a}^{C}_{X})$, and assume that entity
$B$'s contributing opinion $\omega^{B}_{X} = (\vec{b}^{B}_{X},
u^{B}_{X}, \vec{a}^{B}_{X})$ is known.

The averaging fission of these two bodies of evidence is denoted as
$\omega^{A}_{X} = \omega^{C\overline{\underline{\diamond}} B}_{X} = \omega^{C}_{X}
\underline{\ominus} \omega^{B}_{X}$, which represents entity $A$'s contributing
opinion. The mathematical expressions for averaging fission is
described below.

\begin{theorem}{\bf The Averaging Fission Operator}\\
\label{theorem:averaging-fission}
\hspace{1ex}\\ Let $\omega^{C}_{X} = \omega^{A\diamond B}_{X}$ be the
fused average opinion of $\omega^{B}_{X}$ and the unknown opinion
$\omega^{A}_{X}$ over the frame $X = \{x_i\;|\; i =1,\cdots,k\}$.  Let
$\omega^{A}_{X} = \omega^{C\overline{\underline{\diamond}} B}_{X} $ be the opinion
such that:
\begin{eqnarray}
\mbox{Case I:}\;\;\; \mbox{For }\;\; u^{C}_{X} \neq 0\;\; \lor\;\;  u^{B}_{X} \neq 0 :
\nonumber \\
\hline
\nonumber \\
\left\{
\begin{array}{ll}
b^{A}_{x_{i}} = b^{C\overline{\underline{\diamond}} B}_{x_{i}}
&= \frac{2b^{C}_{x_{i}}u^{B}_{X} - b^{B}_{x_{i}}u^{C}_{X}}
{2u^{B}_{X} - u^{C}_{X}}\\\\
u^{A}_{X} = u^{C\overline{\underline{\diamond}} B}_{X}     
&=  \frac{u^{B}_{X}u^{C}_{X}}{2u^{B}_{X} - u^{C}_{X}}
\end{array}
\right.
\label{eq:averaging-fission}
\end{eqnarray}

\begin{eqnarray}
\mbox{Case II:}\;\;\;\mbox{For }\;\; u^{C}_{X} = 0\;\; \land\;\; u^{B}_{X} = 0:
\nonumber \\
\hline
\nonumber \\
\left\{
\begin{array}{ll}
b^{A}_{x_{i}} = b^{C\overline{\underline{\diamond}} B}_{x_{i}}&= \gamma^{B} \, b^{C}_{x_{i}} - \gamma^{C} b^{B}_{x_{i}}\\\\
u^{A}_{X} = u^{C\overline{\underline{\diamond}} B}_{X} &= 0
\end{array}
\right.
\label{eq:zero-div-averaging-fission}\\
\nonumber\\
\nonumber
\mbox{where }
\left\{
\begin{array}{l}
\gamma^{B} = \lim \limits_{\substack{ 
{u^{C}_{X} \rightarrow 0}\\
{u^{B}_{X} \rightarrow 0}
}}
\;\;\frac{2u^{B}_{X}}{2u^{B}_{X} - u^{C}_{X}}\\
\gamma^{C} = \lim \limits_{\substack{ 
{u^{C}_{X} \rightarrow 0}\\
{u^{B}_{X} \rightarrow 0}
}}
\;\;\frac{u^{C}_{X}}{2u^{B}_{X} - u^{C}_{X}}\\
\end{array}
\right.
\end{eqnarray}

Then $\omega^{C\overline{\underline{\diamond}} B}_{X}$ is called the
average fissioned opinion of $\omega^{C}_{X}$ and
$\omega^{B}_{X}$, representing the result of eliminating the opinions
of $B$ from that of $C$. By using the symbol `$\ominus$' to designate
this belief operator, we define $\omega^{C\overline{\underline{\diamond}} B}_{X}\equiv \omega^{C}_{X} \underline{\ominus} \omega^{B}_{X}$.
\end{theorem}

Averaging fission is the inverse of averaging fusion. Its proof and
derivation is based on rearranging the mathematical expressions of
Theorem~\ref{theorem:averaging-fusion}

It can be verified that the averaging fission operator is idempotent, non-commutative and non-associative.

\section{Examples}

\subsection{Simple Belief Fission}

\begin{figure}[h]
\begin{center}
\includegraphics[scale=0.6,angle=-90]{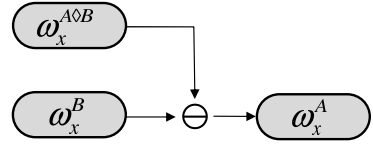}
\caption{Principle of belief fission}
\label{fig:simple-fission}
\end{center}
\end{figure}

Assume that $A$ has an unknown opinion about $x$. Let $B$'s opinion
and the cumulatively fused opinion between $A$'s and $B$'s opinions be
know as:
\[
\begin{array}{l}
\omega^{A\diamond B}_{x} = (0.90,\;0.05,\;0.05,\;\frac{1}{2})\\\\
\omega^{B}_{x}=(0.70,\;0.10,\;0.20,\;\frac{1}{2}) \mbox{and}
\end{array}
\]
respectively. Using the cumulative fission operator it is possible to
derive $A$'s opinion. This situation is illustrated in
Fig.\ref{fig:simple-fission}.

By inserting the opinions values into Eq.(\ref{eq:cumulative-fusion})
the contributing opinion from $A$ can be derived as
\[
\omega^{A}_{x} =
(0.93,\;0.03,\;0.06,\;\frac{1}{2})
\]

\subsection{Inverse Reasoning in Bayesian Networks}

\begin{figure}[h]
\begin{center}
\includegraphics[scale=0.6,angle=-90]{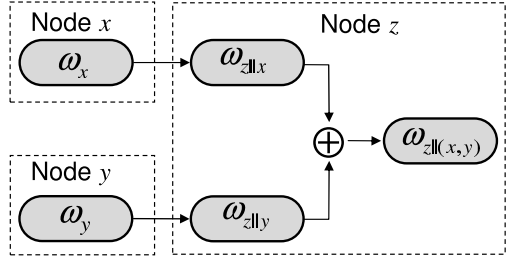}
\caption{Bayesian network with belief fusion}
\label{fig:Bayesian-fusfis}
\end{center}
\end{figure}

Bayesian belief networks represent models of conditional relationships
between propositions of interest. Subjective logic provides operators
for conditional deduction \cite{JPD2005-ECSQARU} and conditional
abduction \cite{PJ2005-ICCRTS} which allows reasoning to take place in
either direction along a conditional
edge. Fig.\ref{fig:Bayesian-fusfis} shows a simple Bayesian belief
network where $x$ and $y$ are parent evidence nodes and $z$ is the
child node.

In order to derive the deduced opinions $\omega_{z\|x}$ and $\omega_{z\|y}$
using the deduction operator, the opinions $\omega_{x}$ and
$\omega_{y}$, as well as the conditional opinions $\omega_{z|x}$,
$\omega_{z|\overline{x}}$, $\omega_{z|y}$ and
$\omega_{z|\overline{y}}$ are needed. Assuming that the contributions
of $\omega_{z\|x}$ and $\omega_{z\|y}$ are independent, they can be
fused with the cumulative fusion operator to produce the derived
opinion $\omega_{z\|(x,y)}$.

Belief revision based on the fission operator can be useful in case a
very certain opinion about $z$ has been determined from other sources,
and it is in conflict with the opinion derived through the Bayesian
network. In that case, the reasoning can be applied in the inverse
direction using the fission operator to revise the opinions about $x$
and $y$ or about the conditional relationships $z|x$ and $z|y$. 

Opinion ownership in the form of a superscript to the opinions is not
expressed in this example. It can be assumed that the analyst derives
input opinion values as a function of evidence collected from
different sources. The origin of the opinions are therefore implicitly
represented as the evidence sources in this model.

\section{Conclusion}

The principle of belief fusion is used in numerous applications. The
opposite principle of belief fission is less commonly used. However,
there are situations where fission can be useful. In this paper we
have described the fission operators corresponding to cumulative and
averaging fusion in subjective logic. The derivation of the fission
operators are based on rearranging the expressions for the
corresponding fusion operators.

\bibliographystyle{plain}
\bibliography{../bibliography}
\end{document}